\documentclass[conference]{IEEEtran}
\IEEEoverridecommandlockouts
\usepackage{cite}
\usepackage{amsmath,amssymb,amsfonts}
\usepackage{algorithmic}
\usepackage{graphicx}
\usepackage{textcomp}
\usepackage{xcolor}
\def\BibTeX{{\rm B\kern-.05em{\sc i\kern-.025em b}\kern-.08em
    T\kern-.1667em\lower.7ex\hbox{E}\kern-.125emX}}
\begin{document}

\title{Simulation-based Analysis Of Highway Trajectory Planning Using High-Order Polynomial For Highly Automated Driving Function\\}

\author{\IEEEauthorblockN{1\textsuperscript{st} Milin Patel}
\IEEEauthorblockA{\textit{Smart Systems} \\
\textit{Hochschule Furtwangen University}\\
Furtwangen, Germany \\
milin.jaileshkumar.patel@hs-furtwangen.de}
\and
\IEEEauthorblockN{2\textsuperscript{nd} Marzana Khatun {3\textsuperscript{rd} Rolf Jung}}
\IEEEauthorblockA{\textit{Electrical Engineering} \\
\textit{Kempten University of Applied Sciences}\\
Kempten, Germany \\
{marzana.khatun; rolf.jung}@hs-kempten.de}
\and
%\IEEEauthorblockN{3\textsuperscript{th} Rolf Jung}
%\IEEEauthorblockA{\textit{Computer Science } \\
%\textit{Kempten University of Applied Sciences}\\
%Kempten, Germany \\
%rolf.jung@hs-kempten.de}
%\and
\IEEEauthorblockN{4\textsuperscript{th} Michael Gla\ss}
\IEEEauthorblockA{\textit{Embedded Systems/Real-Time Systems} \\
\textit{Ulm University}\\
Ulm, Germany \\
michael.glass@uni-ulm.de}

}

\maketitle

\begin{abstract}
One of the fundamental tasks of autonomous driving is safe trajectory planning, the task of deciding where the vehicle needs to drive, while avoiding obstacles, obeying safety rules, and respecting the fundamental limits of road. Real-world application of such a method involves consideration of surrounding environment conditions and movements such as Lane Change, collision avoidance, and lane merge. The main focus of the paper is to develop and implement safe collision free highway Lane Change trajectory using high order polynomial for Highly Automated Driving Function (HADF). Planning is often consider as a higher level process than control. Behavior Planning Module (BPM) is designed that plans the high-level driving actions like Lane Change maneuver to safely achieve the functionality of transverse guidance ensuring vehicle safety and efficient motion planning through environment. Based on the recommendation received from the (BPM), the function will generate a  desire corresponding trajectory. The proposed planning system is situation specific with polynomial based algorithm for same direction two lane highway scenario. The polynomial curve has the advantage of continuous curvature and simplicity that reduces overall complexity and thereby allows rapid computation. The proposed design is verified and analyzed through the MATLAB simulation environment. The results show that the method proposed in this paper has achieved a significant improvement in safety and stability of Lane Changing maneuver.
\end{abstract}


\section{Introduction}
Even though comprehensive research on Highly Automated Vehicles, highway Lane Change problem draws more and more attentions and gave a critical review of the recent design of lane changing models \cite{b1}. The implementation of an autonomous Lane Change system is the main theme of this paper. In the context of automated Lane Change maneuver, trajectory planning is a significant aspect that required remarkable research \cite{Gasparetto}, \cite{Chen} and \cite{Osa}. The main reason for this is the complexity of the Lane Change maneuver, since it incorporates lateral and longitudinal control in the presence of obstacles. The prime focus is to achieve a trajectory from the current lane to the target lane that meets certain constraints, such as acceleration, safety gap, collision avoidance requirements, or other dynamic constraints.

In terms of safety, a computed path must be assured to be collision-free and kinematics feasibility under certain constraints. The focus of the Lane Change planner is operation on highways that involves Lane Change, overtaking and small lateral movements within a lane and overtaking. The focus of this study is limited to trajectory planning in the highway scenario for autonomous Lane Change maneuver. The purpose of the work is to tackle the problem of autonomous Lane Change driving in an uncertain highway environment where the vehicle has to anticipate and adapt to behavior of the surrounding vehicles. Safety is assured by the simulation of the safety-critical areas around the surrounding vehicles that should not be reached by the autonomous vehicle in order to plan an evasive action. 

One of the factors that makes safe trajectory planning difficult is the speed with which the situation around a vehicle can evolve with time.
To compensate, the Lane Change behavior planner often need to make decisions quickly, to respond to ever-changing stimuli in the environment.
This also means the trajectory generation module should generate an optimal trajectory according to the decision made in the behavior planner module.
%Based on the recommendation received from the BPM, the Lane Change trajectory generation module will generate a corresponding desired trajectory and later used by the Model Predictive Control (MPC) based controller as a reference for the vehicle to follow.
Based on the recommendation received from the BPM, the Lane Change trajectory generation module will generate a corresponding desired trajectory and later applying the Model Predictive Control (MPC) based controller a reference path is provided for the vehicle\cite{MatlabWorks}.

In the case of a vehicle is encountered in the adjacent lane during the Lane Change maneuver, the optimal Lane Change trajectory designed when there's no obstacle might not be applied because collision might occur. To meet the criteria of collision free trajectory by increasing the order of polynomial, and the boundary conditions are assumed, then the shape of the path changes. Thus more freedom is provided to avoid the obstacle. A fifth order polynomial can be increased to a sixth order polynomial since the longitudinal direction has has more freedom while lane changing \cite{b2}. Finally, the ideal Lane Change trajectory is planned and set as a desirable trajectory for the ego vehicle to follow, with vehicle dynamic constraints has been applied.

\section{Highway Lane Change Maneuver}
\label{Highway Lane Change Maneuver}
\subsection{Modeled Scenario for Lane Change}
\begin{figure}[htbp]
\centerline{\includegraphics[width=8cm]{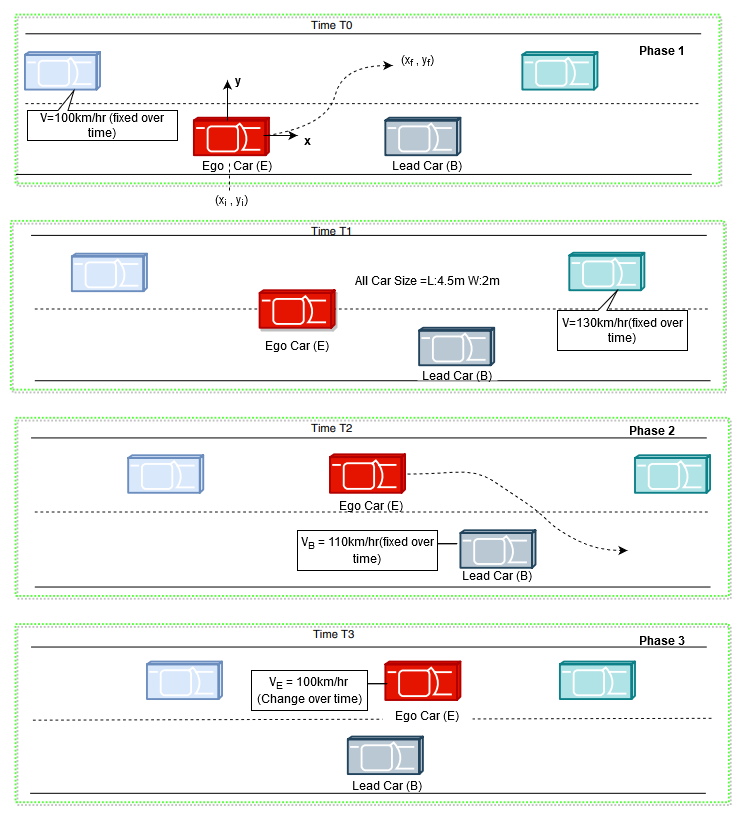}}
\caption{Highway Lane Change scenario\cite{b3}}
\label{fig1}
\end{figure}

Typical lane-change scenario is displayed in Fig.~\ref{fig1} with ego vehicle and other road traffic as described in \cite{b3}. However, scenario modeled parameters are not explicitly mentioned in \cite{b3} and \cite{Pegasus}. The ego vehicle (E) is intended to perform a lane-change maneuver to the adjacent lane. During the movement, two surrounding obstacle vehicles should be considered by the trajectory planning algorithm: the leading vehicle (B) in the same lane as ego vehicle, and two nearest vehicles in the target lane. In order to model the maneuver scenario, the left lane is supposed to be free so that the ego vehicle can overtake without considering additional obstacles.

The change of lateral displacement \textbf{y} along with the change of longitudinal displacement \textbf{x} is the basis of a Lane Change trajectory. Designing a Lane Change trajectory model implies estimating the vehicle's path between two points. The model does not explicitly take into consideration the dynamics of the vehicle or vehicle model for simplicity and generality.

According to author Shamir  \cite{b4}, the Lane Change maneuver is divided into three phases and demonstrated below:
%The Lane Change maneuver is accomplished in three phases:
\begin{itemize}
\item Phase 1: Lane Change to overtake preceding vehicle
\item Phase 2: Navigate straight in adjacent lane
\item Phase 3: Lane Change to return to original lane
\end{itemize}
Phase 1 is examined to demonstrate that the lane-change trajectory's absolute shape, size, and time are unaffected by the obstacle's velocity. For Concrete Scenario modeling, the ego-vehicle as vehicle E, and to the preceding Lead-Car as vehicle B are considered. The initial velocity is of ego vehicle is selected as 27.778 ${m/s^2}$ (100 km/hr) and lead vehicle velocity is 30.558 ${m/s^2}$ (110 km/hr). 

\subsection{Lane Change Behavior Planner}
A behavior planning system plans the set of high-level driving actions for achieving the driving mission under various driving situations. The set of maneuver consider the rules of the road, and the interactions with all static and/or dynamic objects in the environment. The set of high-level decisions are made by the planner that must ensure vehicle safety and efficient motion through the environment. The behavior planner needs to make the right decisions to keep the ego vehicle moving safely towards the goal. With all the necessary information available, the behavior planner must select the appropriate behavior, and define the necessary accompanying constraints to keep the vehicle safe and moving efficiently. 

Author Goswami introduce a set of factors needed to be considered while making decisions by the behavior planner are:\cite{b5}
\begin{enumerate}
\item Distance between the vehicles in the target lane.
\item Velocities of the lead and lag vehicles in the target lane.
\item Velocity of the vehicle in the current lane.
\item Distance to the target vehicle in the current lane.
\item Time taken for the Lane Change maneuver.
\item Distance covered by the ego vehicle during the Lane Change.
\end{enumerate}

The majority of lane-change decision-making is based on the target vehicles' velocities in the current and target lanes, as well as the existence of a gap. Despite the fact that the target vehicles are driving at the uniform speed, the gap between them may decrease or increase depending on the relative velocity\cite{b5}. In this paper Lane Change under accelerated mode have been considered.

\begin{figure}[htbp]
\centerline{\includegraphics[width=8cm]{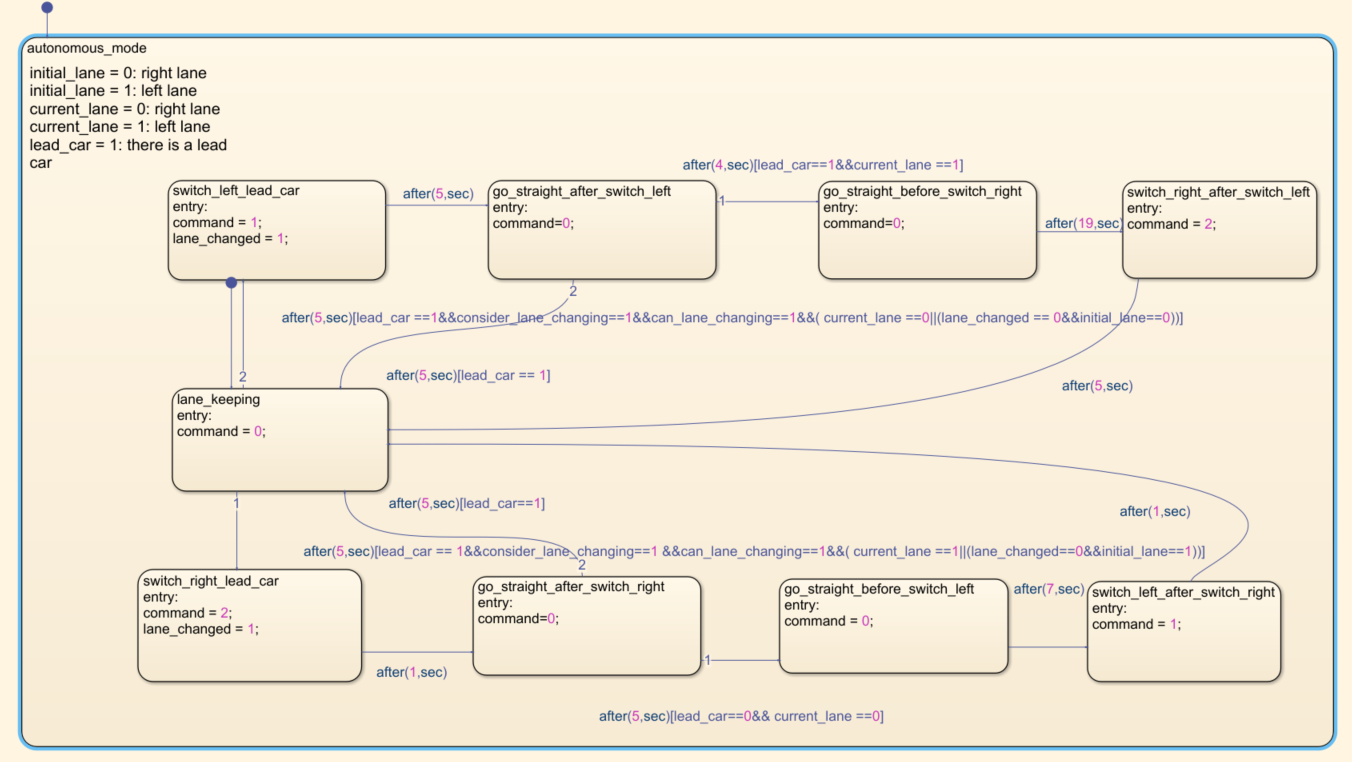}}
\caption{Behaviour Planner for Lane Change maneuver decisions}
\label{fig2}
\end{figure}

An autonomous mode is planned for the Lane Change maneuver that illustrated in Fig.\ref{fig2} representing the behavior planner with initial lane, current lane and target lane for the Lane Change maneuver decision flow diagram. 

\section{Trajectory Generation for Lane Change Scenario}
\subsection{Trajectory Planning}
During the motion phase, trajectory planning is concerned with determining the vehicle's position, velocity, and acceleration. Dealing with polynomials with independent coefficients is a typical method for creating trajectories\cite{b6}. Polynomial curves are computationally simple and contain closed-form formulas that enable continuous curvatures, as observed by Nelson \cite{Nelson} and  Wonshik\cite{b7}.
In order to facilitate, the accelerations at the initial and final points of the lane changing maneuver are presumed to be zero\cite{b4}. During the maneuver itself, the velocity and acceleration are assumed and not constant while driving. In the Lane Change trajectory generation module \cite{Ding}, a geometric high order polynomial trajectory representation is used and the degree of a polynomial can be arbitrary value, but the complexity and calculated amount is becoming more complex with the increase of degree of a polynomial. Three trajectories are formulated using 4$^{th}$, 5$^{th}$ and 6$^{th}$ degrees of polynomial in longitudinal direction to evaluate the idea that suggested trajectories perform motions with continuous velocity and smooth acceleration. First, a longitudinal trajectory is planned and then the lateral motion is planned for a given longitudinal trajectory with regard to safety constraints. 

Nelson \cite{Nelson} suggested as a 5$^{th}$ order polynomial to characterize the lateral position of a vehicle as a function of its longitudinal position for Lane Change maneuvers. In the longitudinal direction, a total of five constraints are obtained (three beginning values of position, velocity, and acceleration) and two final values of (velocity and acceleration) whilst there are six constraints in the lateral direction, as (i) ego vehicle position, (ii) ego vehicle velocity, (iii) ego vehicle acceleration (iv) target vehicle position, (v) target vehicle velocity, and (vi) target vehicle acceleration are specified. These define fourth and 5$^{th}$ polynomial curves for the longitudinal and lateral positions, respectively. The origin of the coordinate frame is fixed in the vehicle's initial position on the right lane with \textbf{x} being the longitudinal axis in the vehicle's driving direction and \textbf{y} being the right-hand side perpendicular to \textbf{y}. 

The equations \eqref{eq1} and \eqref{eq2} for the longitudinal and lateral positions with respect to time (t) are as follows: The equation below is referred from\cite{b7}.
 \begin{equation}
x(t)=a_{0}+a_{1}t+a_{2}t^2+a_{3}t^3+a_{4}t^4
\label{eq1}
\end{equation}
\begin{equation}
y(t)=b_{0}+b_{1}t+b_{2}t^2+b_{3}t^3+b_{4}t^4+b_{5}t^5
\label{eq2}
\end{equation}

However, the variables have been amended according the designed highway scenario. {($a_0$,...$a_4$) and ($b_0$,...$b_5$)} are the coefficients that must be determined for their respective longitudinal and lateral positions using the initial and final constraints.

\subsection{Higher Order Polynomial for Collision Free Trajectory Generation}\label{AA}
The trajectory is altered by increasing the order of one of the polynomials, which is beneficial for enabling more freedom to generate a collision free trajectory\cite{b2}. The challenge is to identify the increase polynomial in order to yield permissible trajectories, and selecting the feasible values of the coefficient. Admittedly, since the vehicle is limited to maneuvering in the current lane and adjacent lane, shaping the lateral trajectory will generate only a few admissible paths, which is not the case in the longitudinal direction. We increase the polynomial degree of the \textbf{x} direction because the changing range of the longitudinal velocity of the vehicle is larger than that of the lateral velocity, often 4 to 5 times larger \cite{b2}. If the order of both $x(t)$ and $y(t)$ are increased, the problem becomes significantly more complex since two additional coefficients shall be calculated. Therefore, increasing the order of $y(t)$ is not a wise choice and increasing the order of $x(t)$ alone is a reasonable option. 

To begin with, fifth-order polynomial function in longitudinal direction can be formulated, with respect to the time (t) can be written as follows: The equations \eqref{eq3}, \eqref{eq4} and \eqref{eq5} for higher order longitudinal trajectory i.e fifth and sixth order are referred from \cite{b2}.
%5th order equations
\begin{equation}
x(t)=a_{0}+a_{1}t+a_{2}t^2+a_{3}t^3+a_{4}t^4+a_{5}t^5
\label{eq3}
\end{equation}
\begin{equation}
\dot x(t)=a_{1}t+3a_{2}t^2+4a_{3}t^3+5a_{4}t^4
\label{eq4}
\end{equation}
\begin{equation}
\ddot x(t)=6a_{3}t+12a_{4}t^2+20a_{5}t^3
\label{eq5}
\end{equation}

Where ${x}$, $\dot{x}$, $\ddot{x}$, ${y}$, $\dot{y}$ and $\ddot{y}$ are the longitudinal displacement, longitudinal velocity, longitudinal acceleration, lateral displacement, lateral velocity and lateral acceleration of the vehicle, respectively. In this way, we simplify the generation of the trajectory of the Lane Change.

Based on the discussion above, the longitudinal trajectory is further modified from a fifth order polynomial to a sixth order polynomial to have more freedom to avoid the obstacle. That means, a 6$^{th}$ degree polynomial in the \textbf{x} direction and 5$^{th}$ degree polynomial in \textbf{y} direction are selected to model the collision avoidance trajectory of the Lane Change. The higher order longitudinal direction trajectory can be represented by:
%6th order equations
\begin{equation}
 x(t)=a_{0}+a_{1}t+a_{2}t^2+a_{3}t^3+a_{4}t^4+a_{5}t^5+6a_{6}t^6
\label{eq6}
\end{equation}
\begin{equation}
 \dot x(t)=a_{1}t+3a_{2}t^2+4a_{3}t^3+5a_{5}t^4+6a_{6}t^4
\label{eq7}
\end{equation}
\begin{equation}
\ddot x(t)=6a_{3}t+12a_{4}t^2+20a_{5}t^3+30a_{6}t^4
\label{eq8}
\end{equation}

The variables and coefficients have been amended according the designed highway scenario as mentioned in section \ref{Highway Lane Change Maneuver} in Fig. \ref{fig1}.

\subsection{Calculation of Trajectory Parameters}
The Lane Change trajectory model based on the quintic polynomial only needs to obtain the initial state and the end state of the vehicle to calculate the Lane Change trajectory cluster \cite{Yue}. According to Lane Change conditions, the following parameters can be known\cite{b8}:
\begin{enumerate}
\item The lateral velocity is zero at the initial position and end position of Lane Change respectively.
\item The acceleration is zero at the initial position and end position of Lane Change respectively.
\item Longitudinal displacement of vehicles at the beginning and end of Lane Change is D, and lateral displacement is ${w}$ (3.6m).
\end {enumerate}

For easier calculation, coefficients should be formulated into a matrix representation.
And based on the boundary condition, further simplification can be made. All coefficients can be represented as function of $d$ and $T$.
%6$^{th}$
The boundary conditions for the longitudinal direction motion are given by as per\cite{b9}:
%\begin{align*}
\begin{gather*}
%\begin{multline*}
x(0) = 0 ,\quad \dot x(0) = v_{i} ,\quad \ddot x(0) = 0 \\ 
 \dot x(T) = v_{t} ,\quad \ddot x(T) = 0
%\end{align*}
\end{gather*}
%\end{multline*}
where $T$ indicates the overall time of the Lane Change maneuver and $v_{i}$ and $v_{t}$ are the initial and the target speed of the vehicle as per the given case scenario. In the lateral direction, for the fifth-order polynomial the boundary conditions for the lateral motion are then set as:\cite{b9}
\begin{gather*}
y(0) = 0 ,\quad \dot y(0) = v_{i} ,\quad \ddot y(0) = 0 \\ 
 \dot y(T) = {w} ,\quad \ddot y(T) = 0
\end{gather*}
where, $w$ represents the lane width. By applying the same boundary condition as mentioned, it is easier to  find the coefficients $a_{i}$ and $b_{j}$ as a function of the constraints.
Then the coefficients for the 4$^{th}$ order longitudinal trajectory and 5$^{th}$ order lateral trajectory are:\cite{b9}
%\begin{gather}
\begin{align*}
\begin{pmatrix}
a_{0} \\ 
a_{1}  \\ 
a_{2}  \\ 
a_{3}  \\ 
a_{4}  \\ 
\end{pmatrix}
=\begin{pmatrix}
{0} \\ 
v_{i}  \\ 
{0} \\ 
\frac{v_{t}-v_{i}}{T^2} \\ 
\frac{v_{i}-v_{t}}{2T^2}  \\ 
\end{pmatrix}
    ;
\begin{pmatrix}
b_{0} \\ 
b_{1}  \\ 
b_{2}  \\ 
b_{3}  \\ 
b_{4}  \\ 
b_{5}
\end{pmatrix}
=\begin{pmatrix}
{0} \\ 
{0} \\ 
{0} \\ 
\frac{10w}{T^4} \\ 
\frac{-15w}{T^4}  \\ 
\frac{6w}{T^5}
\end{pmatrix}
%\end{gather}
\end{align*}

Based on the above discussions, the longitudinal direction trajectory should be increased to higher order, so the coefficients for 5$^{th}$ and 6$^{th}$ order longitudinal polynomial can be represented as as per\cite{b10}:
%\begin{gather}
\begin{align*}
\begin{pmatrix}
a_{0} \\ 
a_{1}  \\ 
a_{2}  \\ 
a_{3}  \\ 
a_{4}  \\
a_{5}  
\end{pmatrix}
=\begin{pmatrix}
{0} \\ 
v_{i}  \\ 
{0} \\ 
\frac{10}{T^3}{(d-{T}v_{i})} \\ 
\frac{-15}{T^4}{(d-{T}v_{i})}\\
\frac{6}{T^5}{(d-{T}v_{i})}
\end{pmatrix}
    ;
 %for the 6th order
\begin{pmatrix}
a_{0} \\ 
a_{1}  \\ 
a_{2}  \\ 
a_{3}  \\ 
a_{4}  \\ 
a_{5}  \\ 
a_{6}  
\end{pmatrix}
=\begin{pmatrix}
{0} \\ 
v_{i}  \\ 
{0}   \\ 
\frac{-1}{T^2}{(v_{i}-v_{t}+a_{6}{T^5})} \\ 
\frac{1}{T^3}({\frac{v_{t}-v_{i}}{T^2}+{3}a_{6}{T^5}})  \\ 
{-3}a_{6}{T^5} \\ 
{0.01}
\end{pmatrix}
\end{align*}

Here, $d$ is total longitudinal distance travelled during the Lane Change maneuver. Now, the extra coefficient $a_{6}$ is free to be designed in order to avoid collision. To decide the extra parameter for the sixth order, a mapping of obstacle and the parameter is applied. Improper selection of $a_{6}$ might result a collision and thus, $a_{6}$ = 0.01 is chosen in the following simulation based on the specification of the given scenario and criteria for mapping of $a_{6}$ is provided in the \cite{b11}. It is obvious that the selection of $a_{6}$ can be part of an optimization analysis for different use case scenarios as only Lane Change highway scenario is considered.

\subsection{Trajectory Optimization}
The optimal trajectory is selected from the lane changing trajectory collected by the polynomial. 
For high enough initial velocities, explicit formulas are obtained for the optimal distance and the optimal time of the maneuver. 
The Lane Change duration can be calculated by:\cite{b9}
\begin{equation}
{T} = \frac{2(D_{r}-D_{s})}{v_{t}+v_{i}-2v_{B}}
\label{eq9}
\end{equation}
Where, $v_{B}$ is the Lead-Car speed (30.558 ${m/s^2}$), $D_{s}$ is the safety distance (set as 3m) and $D_{r}$ is the relative distance. Similarly, the total longitudinal distance traveled during the Lane Change maneuver i.e. $d$ can be calculated using:\cite{b9}
\begin{equation}
{d} = \frac{T}{2}{(v_{t}+v_{i})}
\label{eq10}
\end{equation}

The vehicle dynamics constraints are fulfilled by minimizing the longitudinal and the lateral accelerations of the vehicle. The boundary limits of the allowable longitudinal acceleration of the vehicle are defined by $a_{min}$ and $a_{max}$. The optimal vehicle acceleration can be represented as\cite{b4}:
\begin{equation}
f_{a}(t)=\sqrt{\ddot x^2 - \ddot y^2}
\label{eq11}
\end{equation}

Thus, the constraint can be defined as $f_{a}(t)$ $<$ $a_{max}$.
The maximum and minimum admissible acceleration constraints used on longitudinal acceleration are set as follows:
\begin{align*}
a_{max} = 2{m/s^2},\quad a_{min} = {-3m/s^2}.
\end{align*}

The acceleration fluctuates a bit specially while making Lane Change and then finalizes around zero as the goal of controller is met\cite{b9}. Jerk is the rate of change of acceleration in a vehicle. That means rapid changes in acceleration or deceleration result in a "jerk”. Here we denote $\ddot{x}$ as longitudinal jerk and $\ddot{y}$ as lateral jerk.
\begin{figure}[htbp]
\centerline{\includegraphics[width=8cm]{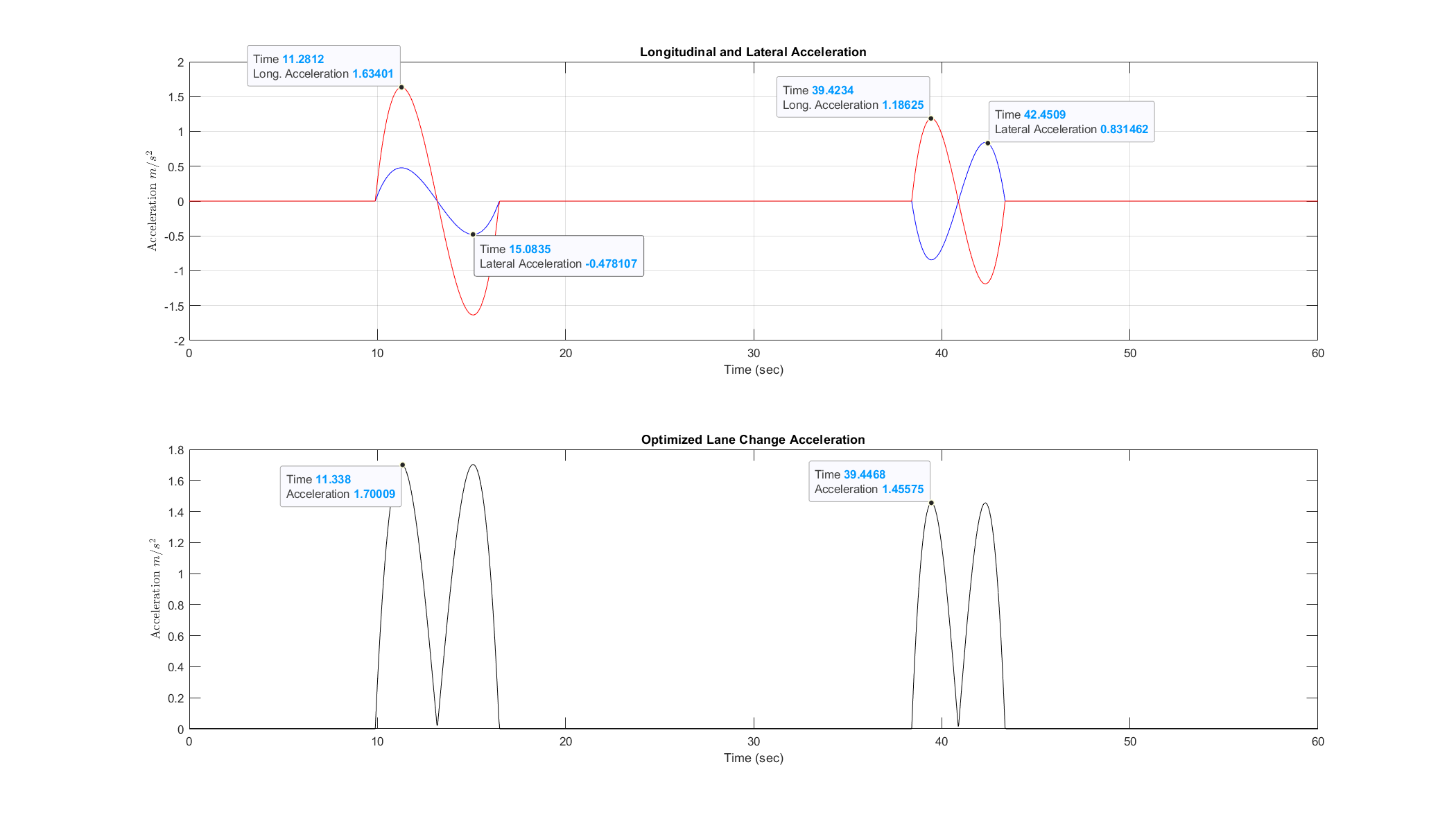}}
\caption{Longitudinal and Lateral Acceleration of the trajectory 5$^{th}$order}
\label{fig3}
\end{figure}

The simulated results of longitudinal and lateral acceleration using \eqref{eq5} and  \eqref{eq11} is illustrated in Fig.~\ref{fig3}. It assures that the motion in the \textbf{x} direction is always forward and vehicle is constantly moving forward or that $\dot x(t)$ $>$ 0. The result validate that the obtained acceleration profile satisfy the acceleration conditions as well as the constraint during a motion considering the Lane Change highway scenario as described in three phase in section \ref{Highway Lane Change Maneuver}.

\begin{figure}[htbp]
\centerline{\includegraphics[width=8cm]{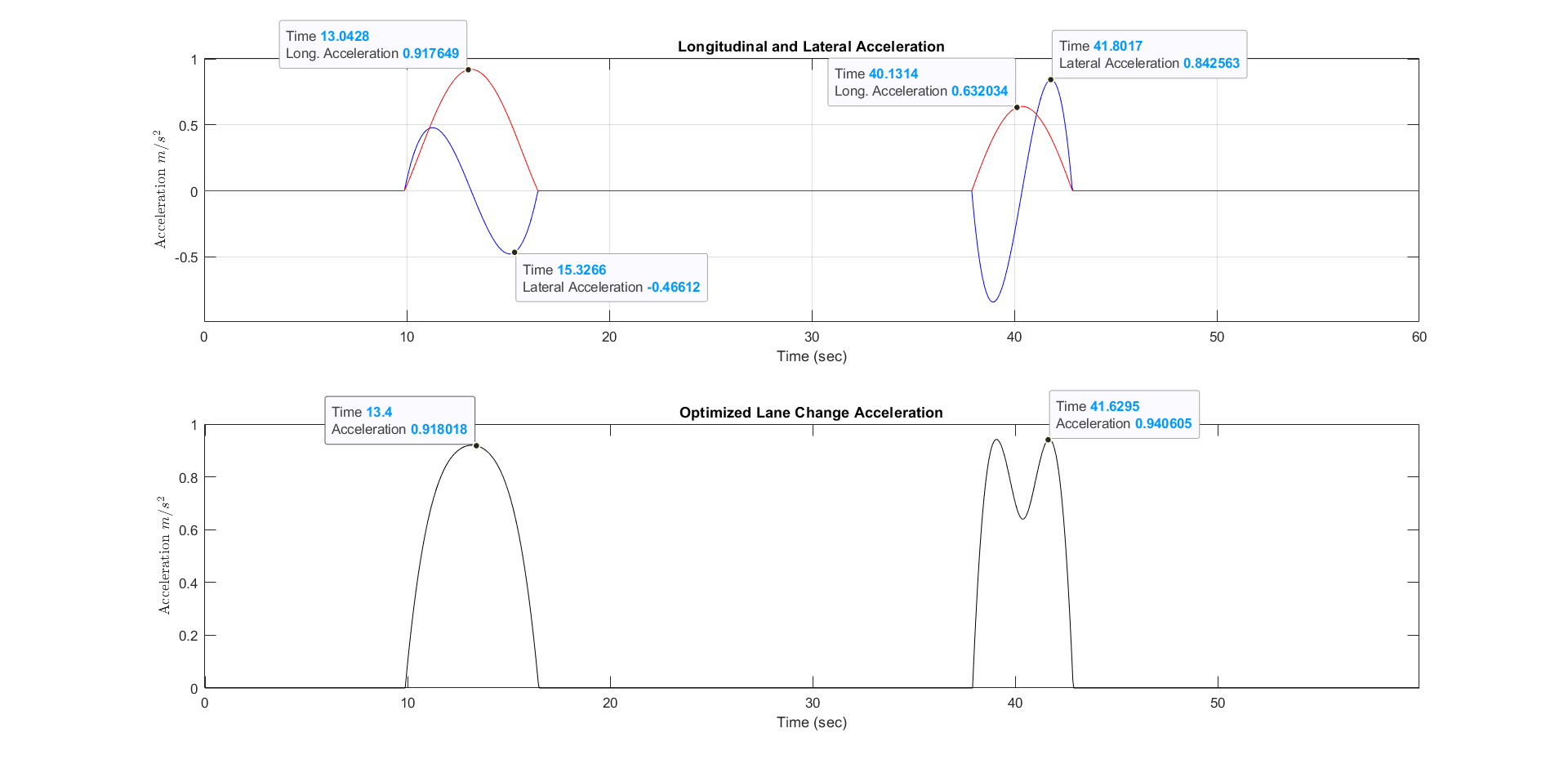}}
\caption{Longitudinal and Lateral acceleration of the trajectory 6$^{th}$order}
\label{fig4}
\end{figure}

The discontinuous acceleration may result in an excessive jerk value. Increasing the order of the polynomial could help solve the problem. With sixth order trajectory, the longitudinal and lateral acceleration of the trajectory using higher order \eqref{eq8} and \eqref{eq11} is illustrated in Fig.~\ref{fig4}.

The plot verifies that adding a new parameter ($a_{6}$) to the longitudinal trajectory has no influence on the lateral trajectory. Thus the Lane Change is considered complete when both longitudinal and lateral accelerations of the controlled vehicle is zero. 

\section{Highway Lane Change Simulation Analysis}
\begin{figure}[htbp]
\centerline{\includegraphics[width=8cm]{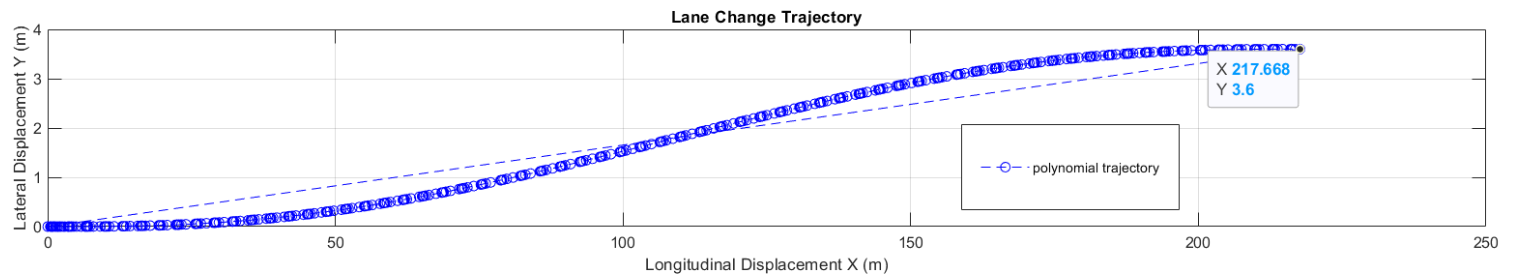}}
\caption{Lane Change trajectory of 5$^{th}$order }
\label{fig5}
\end{figure}

\begin{figure}[htbp]
\centerline{\includegraphics[width=8cm]{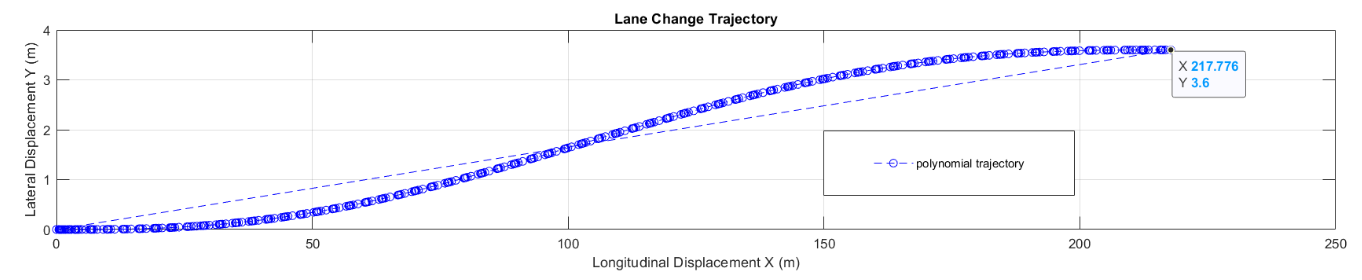}}
\caption{Lane Change trajectory of 6$^{th}$order}
\label{fig6}
\end{figure}

In order to verify the conceivably and effectiveness of the proposed algorithms, the modeling and simulation test are carried out in MATLAB/Simulink. The lateral and longitudinal displacement curves, as well as other associated parameters, are visualized and trajectory is plotted. The following graphs (Fig.~\ref{fig5} and Fig.~\ref{fig6}) show the generated polynomial trajectories of 5$^{th}$ and 6$^{th}$ order respectively. The generated trajectory has a continuous curvature and a simple closed-form.

%5th order Lane Change trajectory
%6th order Lane Change trajectory

The 6$^{th}$ order gives a smooth transition curve and so it is more suitable for avoiding collision specially while performing Lane Change with high velocity and thus under accelerated mode and also covers more distance with the appropriate selection of ($a_{6}$) to the longitudinal trajectory.

%Figures of Lateral and Longitudinal Displacement V/S time.

When the vehicle's lane changing conditions are met, the vehicle's lateral coordinate will increase with time. After the Lane Change is completed, the vehicle's lateral displacement will be lane width ($w$ = 3.6m). The simulation results demonstrate that trajectory and acceleration profiles generated by the proposed model are smooth and continuous and the vehicle can avoid potential collisions efficiently during the lane changing process.

\begin{figure}[htbp]
\centerline{\includegraphics[width=8cm]{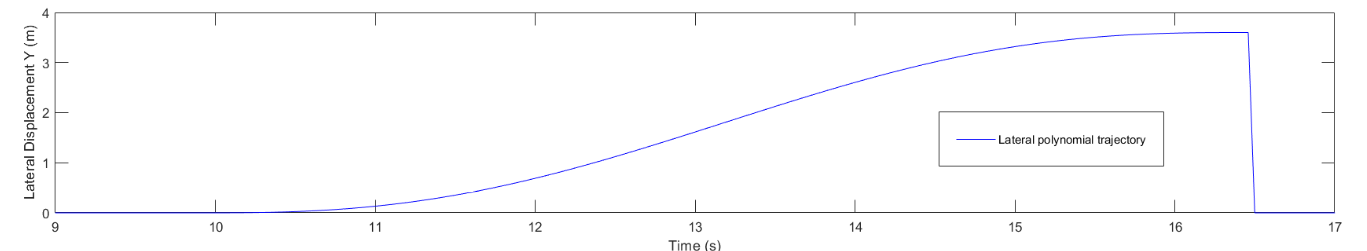}}
\caption{Lateral displacement trajectory of 6$^{th}$order }
\label{fig7}
\end{figure}

As shown in Fig.~\ref{fig7}, the horizontal coordinate of the curve is time, and the vertical coordinate is the vehicle's lateral displacement distance. 

\begin{figure}[htbp]
\centerline{\includegraphics[width=8cm]{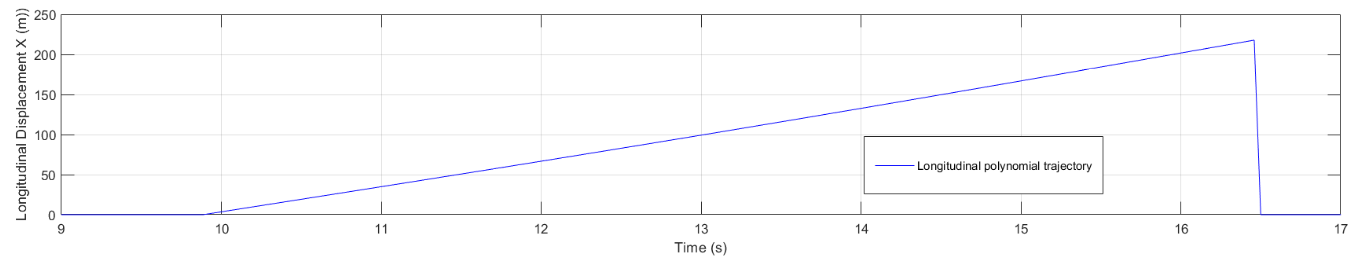}}
\caption{Longitudinal displacement trajectory of 6$^{th}$order}
\label{fig8}
\end{figure}

As demonstrated in Fig.~\ref{fig8}, the longitudinal displacement of the vehicle changes proportionally with time during the lane changing process.

To extend the capability of the proposed algorithm and to ensure realistic behavior,lateral trajectory is modified from fifth order to seventh order polynomial in the lateral direction to avoid collision. In real-world driving scenarios, the resulting trajectories of high order polynomial are very similar.

7$^{th}$ order polynomial Lane Change trajectory equations and calculation are referred from \cite{b12}.
\begin{equation}
y(t)=b_{0}+b_{1}t+b_{2}t^2+b_{3}t^3+b_{4}t^4+b_{5}t^5+b_{5}t^6+b_{5}t^7
\label{eq13}
\end{equation}
Then the coefficients for the 7$^{th}$ order lateral trajectory can be represented as:  \cite{b12}.
\begin{align*}
\begin{pmatrix}
b_{0} \\ 
b_{1}  \\ 
b_{2}  \\ 
b_{3}  \\ 
b_{4}  \\ 
b_{5}   \\
b_{6}   \\
b_{7}
\end{pmatrix}
=\begin{pmatrix}
{0} \\ 
{0} \\ 
{0} \\ 
{0} \\ 
\frac{10w}{T^4} \\ 
\frac{-84w}{T^5}  \\ 
\frac{70w}{T^6} \\ 
\frac{-20w}{T^7}
\end{pmatrix}
\end{align*}

\begin{figure}[htbp]
\centerline{\includegraphics[width=8cm]{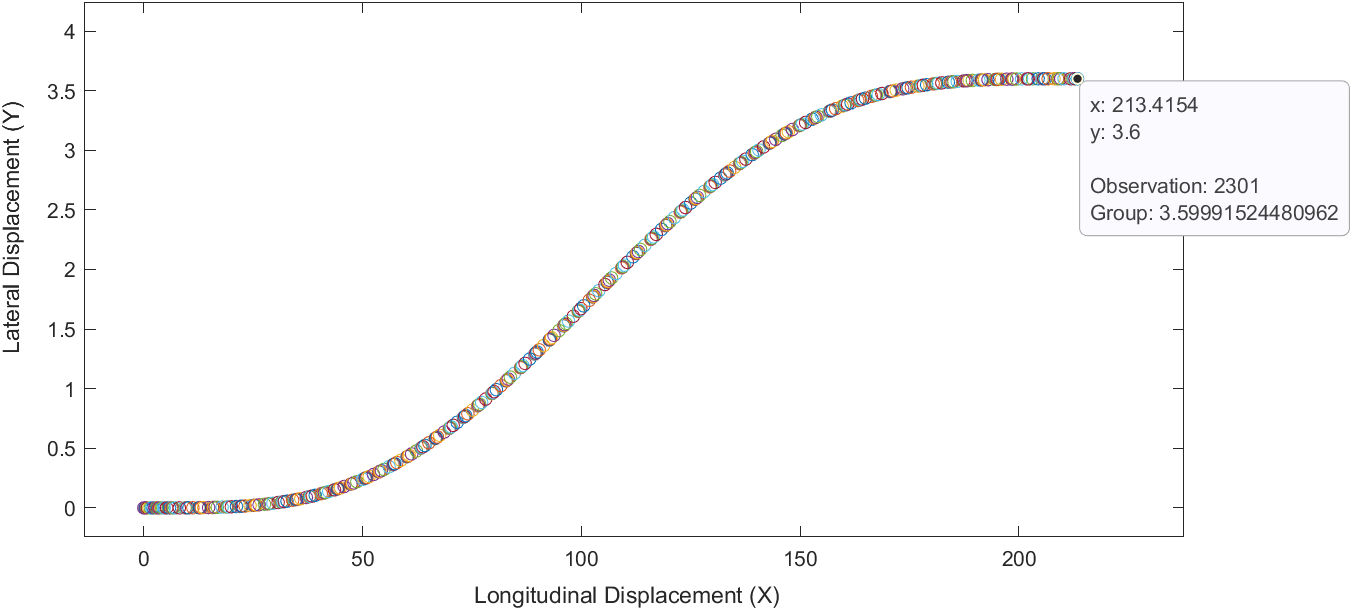}}
\caption{Lane Change trajectory of 7$^{th}$order polynomials in lateral direction}
\label{fig9}
\end{figure}
It can be seen from Fig.~\ref{fig9} that the generated trajectory have a continuous curvature outline and is relatively smooth. In terms of time computations and trajectory stability, the technique based on quintic polynomials revealed more interesting performance of trajectory generation with increasing velocity. We may deduce that during the entire lane changing process, the ego vehicle did not crash with any other vehicles.

\section{Conclusion}
To enhance the validation of the autonomous vehicle safety simulation-based approach is required and has become a hot topic in recent years. In this paper, a lane changing trajectory is planned for lance changing of an autonomous vehicle such as fully automated vehicle. The general lane changing trajectory generation problem were presented and the results of the proposed algorithms are validated in a simulation environment (MATLAB/Simulink). This study proposes the simulation based analysis for trajectory planning problem denoted by 4$^{th}$, 5$^{th}$ and 6$^{th}$ degrees of polynomials. In order to improve the smoothness and accuracy of lane change control, a 7$^{th}$ order polynomial lane change trajectory in lateral direction was implemented. The formulation ensures that during the generated trajectories, the vehicle moves according to abounded acceleration and bounded constraints in order to meet the lane changing requirements safely. The content of the study in this paper is single Lane Change environment, and does not consider complex working conditions and vehicle dynamics models. However, polynomial based algorithm is capable and still finds a safe and efficient lane-change trajectory in high speed environment. The implemented algorithm is a very promising trajectory planning approach that can be beneficial to deal with the actual situation around the vehicle. The comparisons and simulation based validation presented in above chapters demonstrate that the suggested trajectories model can be used in a high-speed environment.

Further research will focus on building vehicle models and road influence factors to verify and improve the algorithm content. In terms of Lane Change trajectory generation, a better method should be investigated, such as an overall optimization process that considers both Lane Change control and Lane Change trajectory, because the method used in this paper generates a trajectory with associated velocity and acceleration, which may not be optimized during the process. Taking into account varying road curvatures is another significant aspect of this work.

\end{document}